\documentclass[conference]{IEEEtran}
\IEEEoverridecommandlockouts
\usepackage{cite}
\usepackage{amsmath,amssymb,amsfonts}
\usepackage{algorithmic}
\usepackage{graphicx}
\usepackage{textcomp}
\usepackage{adjustbox}
\usepackage[numbers]{natbib}
\usepackage[table]{xcolor}
\usepackage{multirow}
\usepackage{url}
\usepackage{subfigure}
\usepackage{subcaption}
\usepackage[utf8]{inputenc}
\usepackage[T5]{fontenc}

\def\BibTeX{{\rm B\kern-.05em{\sc i\kern-.025em b}\kern-.08em
    T\kern-.1667em\lower.7ex\hbox{E}\kern-.125emX}}
\begin{document}

\title{Comparative Opinion Mining in Product Reviews: Multi-perspective Prompt-based Learning\\
}

\author{\IEEEauthorblockN{Hai-Yen Thi Nguyen}
\IEEEauthorblockA{\textit{Faculty of Information Technology} \\
\textit{VNU - University of Engineering and Technology}\\
Hanoi, Vietnam \\
20021480@vnu.edu.vn}
\and
\IEEEauthorblockN{Cam-Van Thi Nguyen$^\dagger$\thanks{$^\dagger$Corresponding author.
Cam-Van Thi Nguyen was funded by the Master, PhD Scholarship Programme of Vingroup Innovation Foundation (VINIF), code VINIF.2023.TS147.}}
\IEEEauthorblockA{\textit{Faculty of Information Technology} \\
\textit{VNU - University of Engineering and Technology}\\
Hanoi, Vietnam \\
vanntc@vnu.edu.vn}
}

\maketitle

\begin{abstract}
Comparative reviews are pivotal in understanding consumer preferences and influencing purchasing decisions. Comparative Quintuple Extraction (COQE) aims to identify five key components in text: the target entity, compared entities, compared aspects, opinions on these aspects, and polarity. Extracting precise comparative information from product reviews is challenging due to nuanced language and sequential task errors in traditional methods. To mitigate these problems, we propose \textbf{MTP-COQE}, an end-to-end model designed for COQE. Leveraging multi-perspective prompt-based learning, MTP-COQE effectively guides the generative model in comparative opinion mining tasks. Evaluation on the Camera-COQE (English) and VCOM (Vietnamese) datasets demonstrates MTP-COQE's efficacy in automating COQE, achieving superior performance with a 1.41\% higher F1 score than the previous baseline models on the English dataset. Additionally, we designed a strategy to limit the generative model's creativity to ensure the output meets expectations. We also performed data augmentation to address data imbalance and to prevent the model from becoming biased towards dominant samples.
\end{abstract}


\section{Introduction}
\label{sec:intro} 
\begin{figure}[t!]
    \centering
    \includegraphics[width=0.95\linewidth]{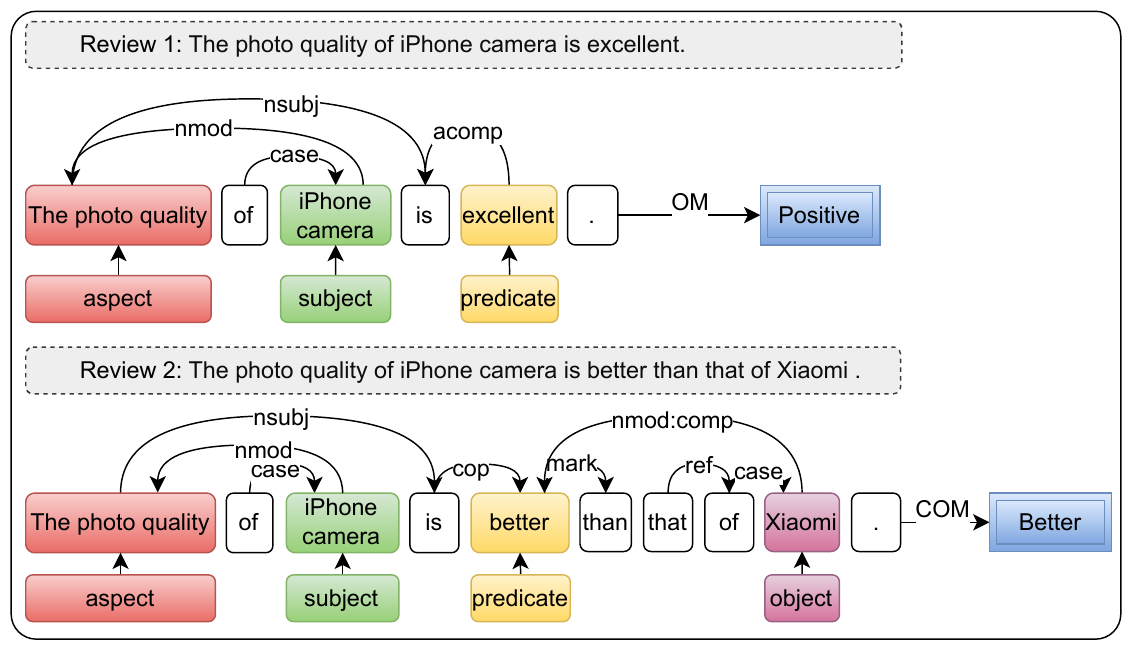}
    \caption{Non-comparative and comparative review examples.}
    \label{fig:review-example}
\end{figure}

The internet hosts a vast array of opinions from individuals across the globe, covering an extensive range of topics and reflecting diverse perspectives and experiences. 
Beyond merely expressing positive or negative sentiments about entities or aspects, consumers frequently engage in comparing similar entities or aspects, generating what are known as comparative opinions. Comparison is essential for evaluating various entities, whether they are products, services, people, actions, or other aspects. 
Traditional opinion mining primarily focuses on determining the polarity (positive, negative, or neutral) of an opinion towards a specific target. However, comparative opinions, where consumers compare two or more entities based on certain aspects, offer richer insights. 
Figure \ref{fig:review-example} illustrates examples of non-comparative and comparative reviews. Unlike non-comparative sentences, comparative sentences necessitate a reference to a comparison target, adding complexity to their analysis.

Numerous studies have focused on individual tasks within comparative sentiment analysis, such as Comparative Sentence Identification (CSI), Comparative Element Extraction (CEE), or Comparative Preference Classification (CPC). \citet{CSI_jindal_2006} were pioneers in introducing comparative opinion mining, identifying CSI and CEE as critical subtasks. CSI determines whether a sentence is comparative, while CEE extracts all comparative elements within such sentences. Subsequently, \citet{panchenko2019categorizing} introduced the CPC task, aiming to classify the comparative preference (BETTER, WORSE, or NONE) within a sentence. To integrate these subtasks, \citet{COQE} proposed the concept of Comparative Quintuple Extraction (COQE), which involves extracting five key pieces of information from a comparative sentence: subject, object, aspect, predicate, and comparison type. They initially employed a multi-stage pipeline approach to address CSI, CEE, and CPC sequentially. While this method simplifies development and allows for focused application of NLP techniques, it is prone to error propagation and suboptimal joint performance.

\begin{figure*}[!t]
    \centering
    \includegraphics[width=0.7\linewidth]{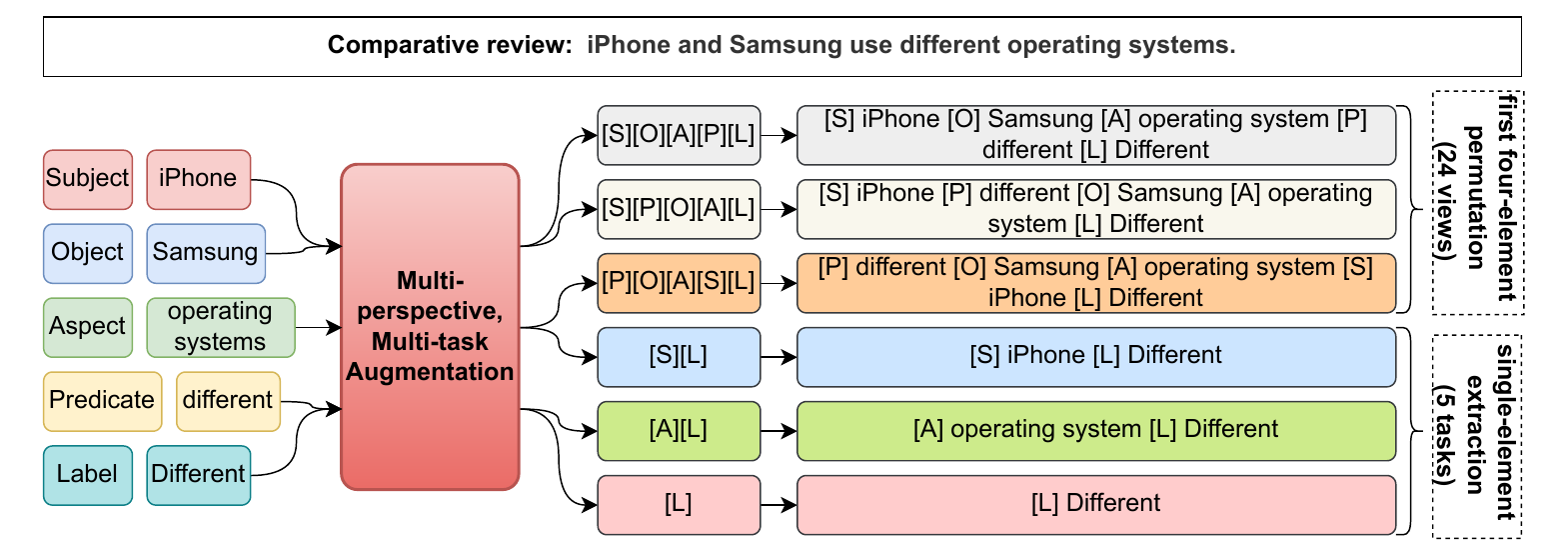}
    \caption{Multi-perspective, Multi-task Learning. Element order permutation for augmenting train dataset size.}
    \label{fig:mtv-mtt-learning}
\end{figure*}
There are two main approaches to solving the quintuple extraction problem: the \textit{pipeline architecture} and the \textit{end-to-end (E2E) architecture}.
The Pipeline Architecture is a common approach where complex tasks are divided into a series of smaller, sequential steps. Each step performs a specific function, and its output feeds into the next step \cite{COQE, le2024unveiling}. However, errors from one stage can propagate through the entire pipeline, potentially leading to inaccurate final results.
In contrast, the End-to-End (E2E) Architecture is a design framework where a system or process handles all necessary steps to complete a task without relying on intermediate components or conversions. This model takes a review sentence as input and directly outputs the quintuple. \citet{gcn-e2e} enhanced neural network representation by incorporating GCN layers with sentence syntactic information. Approaches such as \cite{COQE, DAP, vanthin2023comom} utilized pre-trained models for COQE, implementing various strategies to enhance performance. Specifically, \citet{UniCOQE} designed a predict-and-assign paradigm and trained the model twice, while \citet{DAP} leveraged large language models (LLMs) for data augmentation, and \citet{vanthin2023comom} enriched training sources through multi-task instruction.

In recent years, generative models have emerged as a powerful category of machine learning models capable of creating entirely new data, such as text, code, or images. For COQE, generative models offer a promising approach to automatically extract comparative opinions from text. However, a significant challenge with generative models, particularly in COQE, is the risk of hallucination—generating plausible but incorrect information that is not supported by the input data.

In this study, we introduce \textbf{MTP-COQE}, a novel unified generative model designed for COQE task. Our approach adopts an end-to-end methodology to address the issue of error propagation inherent in traditional pipeline architectures. MTP-COQE utilizes multi-perspective prompt-based learning to effectively guide the generative model in comparative opinion mining tasks. We evaluate MTP-COQE on two diverse datasets: Camera-COQE (English) \cite{COQE} and VCOM (Vietnamese) \cite{hoang2023overview}, demonstrating its efficacy in automating COQE by extracting comprehensive quintuples in a cohesive process. To boost performance, we integrate pre-trained generative models and propose a unique augmentation strategy to enrich training data. Additionally, we implement a controlled creativity approach to align the model's outputs closely with expected results; and also addresses data imbalance through extensive data augmentation techniques, ensuring robustness against biases towards dominant samples in the dataset.

\section{Methodology}
\label{sec:method}
Given an input sentence, the primary goal of \textbf{MTP-COQE} is to detect comparisons within a single review sentence and extract comparative quintuples in a unified process. \textbf{A comparative quintuple} is structured as follows:
\begin{equation}
    Q_5 = (S, O, A, P, L)
\end{equation}
\begin{itemize}
\item \textbf{Subject $S$}: Refers to the entity under evaluation, such as a specific product, model within a product line, or comparable entity.
\item \textbf{Object $O$}: Represents the entity against which the subject is compared, serving as the reference point.
\item \textbf{Aspect $A$}: Describes the specific feature or attribute being compared.
\item \textbf{Predicate $P$}: Expresses the comparative relationship between the subject and object.
\item \textbf{Label $L$}: Indicates the type of comparison made, categorizing comparisons into:
\begin{itemize}
\item Gradable comparisons (e.g., ``better'', ``smaller'').
\item Superlative comparisons (e.g., ``best'', ``smallest'').
\item Equal comparisons (e.g., ``same'', ``as good as'').
\item Non-gradable comparisons (e.g., ``different from'', ``unlike'').
\end{itemize}
\end{itemize}

Our model \textbf{MTP-COQE} consists of three main components: (1) \textit{multi-perspective augmentation} (Figure \ref{fig:mtv-mtt-learning}), (2) \textit{tranfer learning with generative prompt template} (Figure \ref{fig:model-transfer}) and (3) \textit{constrained decoding}. 

\subsection{Multi-perspective Augmentation}
We leverage the multi-view concept \cite{gou2023mvp} and propose a multi-perspective, multi-task learning approach to enhance training data. The permutation strategy is depicted in Figure \ref{fig:mtv-mtt-learning}. Within the quintuple's five elements, only the label $L$ requires classification, while the remaining four elements represent extracted words from the input sequence. Thus, to differentiate between the tasks of CEE and CPC, we permute only the order of the first four elements $(S, O, A, P)$ while keeping the last element $L$ fixed. Additionally, we augment the dataset with single extraction tasks to help it better understand each component of the quintuple.

For non-comparative sentences, permuting elements within the quintuple is meaningless; therefore, we only apply data augmentation to comparative sentences. This also addresses the issue of insufficient labeled data in the training dataset. Augmented data replaces the original dataset and is utilized for fine-tuning the pre-trained model through transfer learning on the COQE task.

\subsection{Transfer Learning with Generative Prompt Template}

\begin{figure}[!t]
    \centering
    \includegraphics[width = \linewidth]{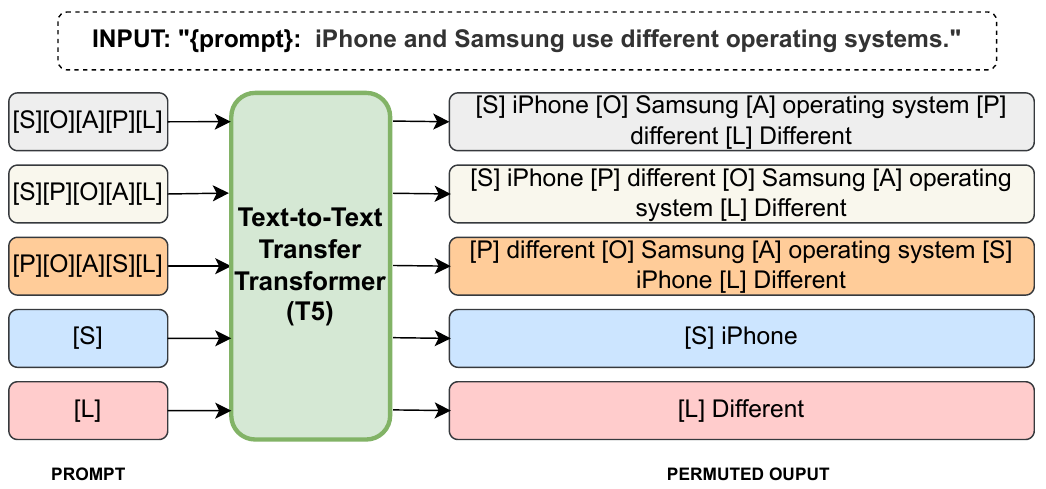}
    \caption{Transfer learning with generative prompt template.}
    \label{fig:model-transfer}
\end{figure}

In MTP-COQE, we employ the Text-to-Text Transfer Transformer (T5) architecture \cite{T5} as the backbone sequence-to-sequence (Seq2Seq) model. We have designed a unified generation template (illustrated in Figure \ref{fig:model-transfer}), which defines a standardized format for input and output to optimize the transfer learning process.

We start with a list of sentences \( S = \{s_1, \ldots, s_m\} \), where \( m \) represents the number of sentences. Each sentence \( s_i \) is tokenized using an appropriate NLP tool (e.g., CoreNLP for English datasets and VnCoreNLP for Vietnamese datasets) to establish positional indices for each token in the input sequence. We utilize markers \( [S] \), \( [O] \), \( [A] \), \( [P] \), and \( [L] \) in the generated output. For each permutation \( p_j \), the corresponding marker is prefixed to each element and concatenated to form the target sequence \( q_{p_{ij}} \). Subsequently, we prepend the task-specific prompt to each input sentence to construct the final input \( s_{p_{ij}} \), where prefix-task prompts are concatenated according to the correct element order. 
    
    
If an element is missed, it is replaced by [UNK] token. If a sentence has multiple quintuples, each quintuple is separated by semicolon ``;''. Each permuted pair $(s, q)$ will serve as a sample $(X, Y)$ in the supervised training dataset, where $X$ is a tokenized sentence with accompanying prefix prompting, and $Y$ is truth quintuple. The goal of transfer learning is to train the language model on the observations $(X, Y)$ to learn how to perform the COQE task.

For each input sentence $X = \left\{x_{i}\right\}_{i=1}^{n}$, $n$ is the length of input sequence, we send $X$ into the T5-encoder to get the latent representation of the sentence:
\begin{equation}
h^{enc}=\operatorname{T5\_Encoder}(X)
\end{equation}
We then used T5-decoder to predict all the comparative quintuples autoregressively. At the $c^{th}$ moment of the decoder, $h^{enc}$ and the previous output tokens: $y_{1: c-1}$ are utilized as the input into the decoder:

\begin{equation}
h_{c}^{dec}=\operatorname{T5\_Decoder}\left(h^{enc}, y_{1: c-1}\right)
\end{equation}

The conditional probability of token $t_{c}$ is defined as follows:

\begin{equation}
p\left(y_{c} \mid X; y_{1: c-1}\right)=\operatorname{Softmax}\left(h_{c}^{dec} W+b\right) 
\end{equation}

where $W \in \mathbb{R}^{d_{h} \times|\mathcal{V}|}, b \in \mathbb{R}^{|\mathcal{V}|}$. $V$ here refers to the vocabulary size of T5, $W$ is the trainable parameter. 

Generation task was trained on $(X, Y)$ pairs from augmented dataset by minimizing the cross-entropy loss:

\begin{equation}
    \mathcal{L}=-\sum_{i=1}^{l} \log p\left(y_{i} \mid {X} ; y_{\leq i-1}\right)
\end{equation}
where $l$ denotes the target sequence length and $y_{\leq i-1}$ is the previously generated tokens. 

\subsection{Constrained Decoding}

A drawback of the generative model is that we cannot control the specific words generated by the model. The model may leverage existing knowledge to generate outputs, which can be limiting in COQE task where precise control over the generated words is required. We employ constrained decoding \cite{cao2021autoregressive} to ensure each generated token of the output sequence is selected from the input sentence or the predefined tagger tokens, in order to prevent invalid output sequences.

After getting trained model, we can directly conduct inference on the test set to extract the comparative tuples $\hat{\mathbf{y}}$. During the inference, we employ constrained decoding: 

\begin{equation}
    \hat{y}_{i}={\arg\max}_{y_{j} \in \mathcal{U}}  p\left(y_{j} \mid {X} ; \hat{y}_{\leq i-1}\right)
\end{equation}

where $\mathcal{U}=\left\{x_{i}\right\}_{i=1}^{n} \cup\left\{\left\langle m_{j}\right\rangle\right\}_{j=1}^{k}$, $\left\{x_{i}\right\}_{i=1}^{n}$ contains $n$ tokens from input $X$ and $\left\{\left\langle m_{j}\right\rangle\right\}_{j=1}^{k}$ are $k$ continuous tokens initialized by embedding of the tagger tokens $\left\{m_{j}\right\}_{j=1}^{k}$ 

Following the acquisition of the generated quintuples, the next step involves identifying the positions of their comparative elements in the original sentence and filtering invalid generated elements. In the index mapping, if multiple possible indexes of $(S,O,A,P,L)$ exist, the index closest to $P$ as the predicate is chosen. Tuple validation involves checking the generated elements, removing words that do not exist in the original one. 

\section{Experimental Settings}
\subsection{Datasets}

We evaluate our proposed model on two benchmark corpora: Camera-COQE \cite{COQE} for the English language and VCOM \cite{hoang2023overview} for the Vietnamese language.

\begin{table}[!ht]
    \centering
    \begin{tabular}{l l|| r|r }
    \hline
        ~ & ~ & \textbf{VCOM} & \textbf{Camera-COQE}  \\ \hline
        \multirow{7}{*}{Sentence} & Sentences & 9225 & 3304\\ 
        & \#Non-com & 7427 & 1599 \\ 
        & \#Com & 1798 & 1705  \\ 
        & \#Multi-com & 319 &  500 \\ 
        & \#Multi-com/\#Com & 27.48\% & 29.3\%  \\ 
        & \#Com/Sentences & 19.49\% &  51.60\%\\
        & \#Quintuples per Sent & 1.37 & 1.4  \\ \hline
        \multirow{5}{*}{Element}
        & Subject entities  & 2169 & 1649  \\ 
        & Object entities & 1455 & 1316  \\ 
        & Aspect entities & 2068 & 1368   \\ 
        & Predicate entities & 2468 & 2163  \\ 
        & Label types & 8 & 4 \\
        \hline
    \end{tabular}
    \caption{Statistics of VCOM  \cite{hoang2023overview} and Camera-COQE \cite{kessler-kuhn-2014-corpus}.}
    \label{dataset-statistic}
\end{table}

Table \ref{dataset-statistic} show the basic statistics about some aspects of VCOM and Camera-COQE.
\begin{figure}[!ht]
    \centering
    \begin{minipage}[b]{0.45\linewidth}
        \centering
        \includegraphics[width=\linewidth]{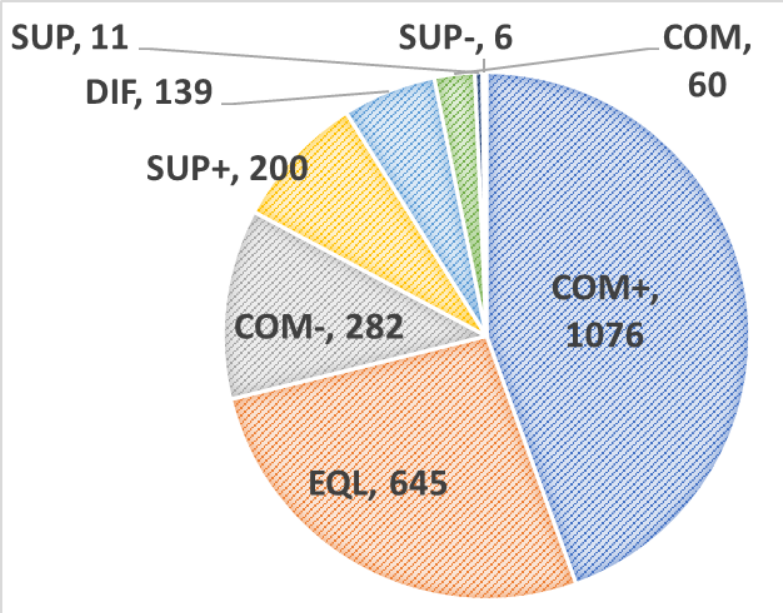}
        \caption{VCOM}
        \label{fig:vcom-label}
    \end{minipage}
    \hfill
    \begin{minipage}[b]{0.45\linewidth}
        \centering
        \includegraphics[width=0.72\linewidth]{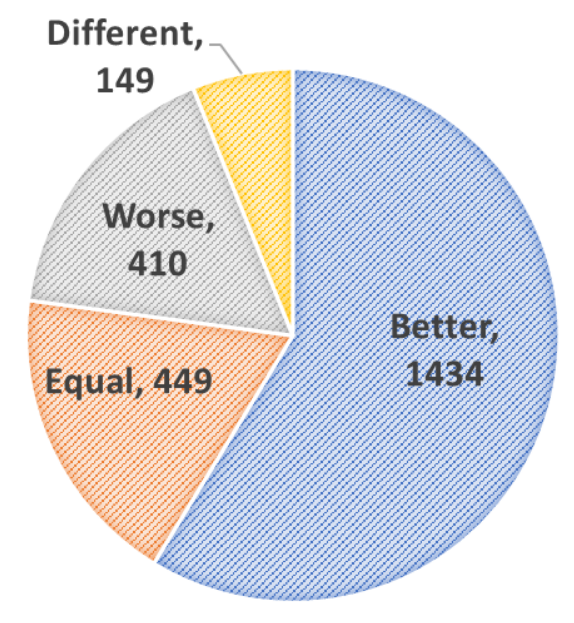}
        \caption{Camera-COQE}
        \label{fig:camera-coqe-label}
    \end{minipage}
\end{figure}
Figure \ref{fig:vcom-label} and \ref{fig:camera-coqe-label} show the distribution of comparison type in the  VCOM and Camera-COQE corpus. This difference in quantity can cause difficulties in training the model, as the model will not have enough data on minority labels and may be biased towards majority labels. 

\subsection{Experiment Scenarios}

For Camera-COQE \cite{COQE}, we utilize the \texttt{google-t5/T5-base} model \cite{T5} as pre-trained backbone model to fine-tuning the downstream COQE task. We conduct experiments with different prompt design strategies to choose the right prompt template giving better performance.

For VCOM \cite{hoang2023overview}, we also re-implement the multi-stage BERT-based model \cite{COQE} (\textbf{MS\textsubscript{BERT}}) on the VCOM corpus to compare the effectiveness of the model when applying a different language dataset. We utilize the \texttt{VietAI/T5-base} model \cite{phan-etal-2022-vit5} to leverage its understanding and processing capabilities of the Vietnamese language. 

\begin{figure}[!ht]
    \centering
    \includegraphics[width =\linewidth]{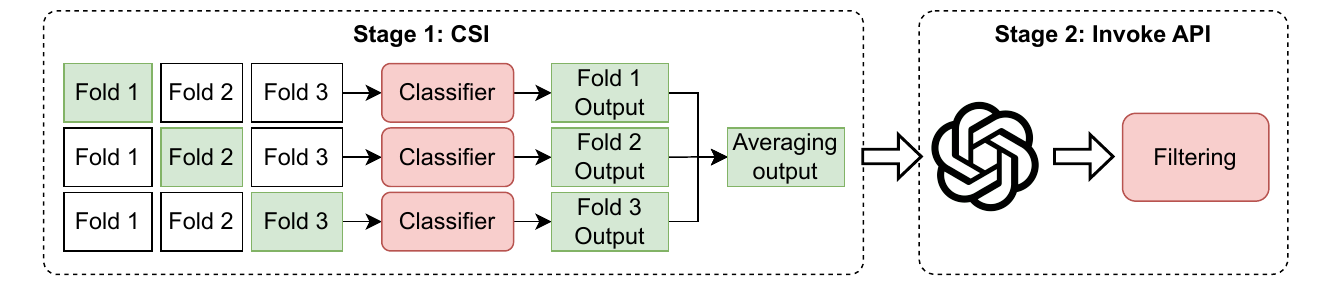}
    \caption{Dual-stage framework for calling \textit{GPT-3.5 Turbo} extracting quintuple.}
    \label{fig:chatgpt-dual-stage}
\end{figure}

Additionally, we use the OpenAI API\footnote{\url{https://openai.com/index/openai-api/}}, \texttt{GPT-3.5 Turbo} to assess the power of large language models (LLMs) in various natural language processing tasks. To minimize token usage when using the OpenAI API, we designed a dual-stage framework (Figure \ref{fig:chatgpt-dual-stage}).

In the ablation experiments, we systematically removed key components of the model. By removing the data augmentation part, we obtained \textbf{T5-with-CD}. Further removing the constraint vocabulary part, we derived \textbf{T5-base} experiment.

\subsection{Baseline Models}
\subsubsection*{Baseline models using Camera-COQE dataset}
\begin{itemize}
    \item \textbf{DAP} \cite{DAP}, where model utilizes external data augmentation (OpenAI API generation).
    \item \textbf{GCN-E2E} \cite{gcn-e2e}, where model enhance embedding representation by adding GCN layers to learning syntactic information.
    \item The models with a multi-stage approach \cite{COQE}, characterized by variations in the main techniques at each stage: \textbf{MS\textsubscript{BERT}}, \textbf{MS\textsubscript{LSTM}}, \textbf{MS\textsubscript{CRF}},
    \textbf{MS\textsubscript{CSR-CRF}}, where CSI task use class sequential rule (CSR) method and CEE task use Conditional Random Field (CRF) method.
    
\end{itemize}

\subsubsection*{Baseline models using VCOM dataset}


\begin{itemize}
    \item The baseline model provided by the VCOM organizers, which are \textbf{VCOM\_pipeline} and \textbf{VCOM\_generative}.
    \item Proposed model by \textbf{ABCD\_team} \cite{vanthin2023comom}, using multi-task instruction tuning, with two versions \textbf{ABCD\_team\textsubscript{viT5-base}} and \textbf{ABCD\_team\textsubscript{viT5-large}}.
    \item Proposed model by \textbf{The challengers} \cite{le2024unveiling}, using sequential framework.
    \item Official results of other participating teams \cite{hoang2023overview}.
\end{itemize}

\subsection{Evaluation Metrics}

We evaluate the performance on CEE, and COQE respectively. For CEE, we use \texttt{Precision}, \texttt{Recall}, \texttt{F1} metrics for each element and their\texttt{ macro-average F1}, \texttt{micro-average F1}. We assess the accuracy of predictions using three matching strategies:
\begin{itemize}
    \item \textbf{Exact Match (E): } The entire extracted element must precisely match the ground truth (Equation \ref{eq:e-match}).

            \begin{equation}\label{eq:e-match}
            \# { correct }_{e} = \left\{\begin{array}{lc}
        0 & \exists\left(g_{k} \neq p_{k}\right) \\
        1 & { otherwise }
        \end{array}\right.
        \end{equation}
        
        where $g_{k}$ denotes $k^{{th }}$ element in the truth comparative quintuple, $p_{k}$ denotes $k^{{th }}$ element in the predicted comparative quintuple. It means that if all $p_{k}$ and $g_k$ match exactly, the count is 1 , otherwise 0 .
    \item \textbf{Proportional Match (P): } We consider the proportion of matched words in the extracted element compared to the ground truth (Equation \ref{eq:p-match}).
            \begin{equation}\label{eq:p-match}
            \# { correct }_{p} = \left\{\begin{array}{lc}
        0  & \exists\left(g_{k} \cap p_{k}=\emptyset\right)\\
        \frac{\sum_{k} {len(g_{k} \cap p_{k})}}{\sum_{k} {len(g_{k})}} & { otherwise }
        \end{array}\right.
        \end{equation}
        where  $len(\cdot)$ denotes the length of the comparative elements.
    \item \textbf{Binary Match (B):} At least one word in the extracted element overlaps with the ground truth (Equation \ref{eq:b-match}).

            \begin{equation}\label{eq:b-match}
            \# { correct }_{b}=\left\{\begin{array}{lc}
        0 & \exists\left(g_{k} \cap p_{k}=\emptyset\right) \\
        1 & { otherwise }
        \end{array}\right.
        \end{equation}
        
        where the count is 1 if all $p_{k}$ and $g_{k}$ have overlaps, otherwise 0.

\end{itemize}

The evaluation of quintuples involves two types of tuples: quadruple (\texttt{Q4}, unconsidered comparison type label) and quintuple (\texttt{Q5}, considered comparison type label).

\section{Results and Analysis}

\subsection{Main Results}
Table \ref{tab:camera-coqe-final-result} presents the results of our experimental scenarios using Camera-COQE corpus compared to the baseline models, based on the $F1$ score. It can be observed that end-to-end methods are achieving higher results compared to pipeline methods. Our model, \textbf{MTP-COQE}, in terms of \texttt{E-Q5-F1} metric, ranks first with a difference of around $9\%$ compared to the highest score pipeline model and $1.41\%$ compared to the second highest score model.

\begin{table}[!ht]
    \centering
    \caption{Performance comparison of different approaches on the Camera-COQE corpus for CSI, CEE tasks, and Q5 under the Exact Matching metric. Model approaches are categorized as \textit{E} for end-to-end, \textit{P} for pipeline, and \textit{H} for hybrid methods. Blue rows indicate our experimental results.}
    \begin{adjustbox}{width=\linewidth}
    \begin{tabular}{c l | r r r }
    \hline
        \textbf{Approach} & \textbf{Method}   & \textbf{Exact-Q5} & \textbf{Binary-Q5} & \textbf{Prop-Q5} \\ \hline
        \rowcolor{blue!10}
        E & MTP-COQE\textsubscript{prefix}  & \textbf{22.47} & \textbf{33.60} & \textbf{31.43} \\ \hline
        E & DAP & 21.06 & \_ & \_\\ \hline
        \rowcolor{blue!10}
        E & MTP-COQE\textsubscript{suffix}   & 19.95 & 31.58 &  29.28\\ \hline
         \rowcolor{blue!10}
        E & T5 with CD   & 18.58 & 26.20 & 24.79\\ \hline
        \rowcolor{blue!10}
        E & T5-base   & 17.89  & 25.26 &23.98\\ \hline 
        E &GCN-E2E & 14.10 & 29.06 & 26.09 \\ \hline
        P & MS\textsubscript{BERT} & 13.36 & 25.25 & 23.26 \\ \hline

        P & MS\textsubscript{LSTM} &  9.05 & \_ & \_ \\ \hline
        P  & MS\textsubscript{CRF}  &  4.88 & \_ & \_ \\ \hline
        P &MS\textsubscript{CSR-CRF}  &  3.46 & \_ & \_ \\ \hline
        \rowcolor{blue!10}
        H & gpt3.5-turbo & 6.60 & 14.35 &13.31 \\ \hline
    \end{tabular}
    \end{adjustbox}
    
    \label{tab:camera-coqe-final-result}
    
\end{table}

For VCOM, we compared our results with teams participating in the VLSP2023 VCOM shared task\footnote{\url{https://vlsp.org.vn/vlsp2023/eval/comon}}, shown in Table \ref{tab:vcom-result}. The E2E structure still has an advantage in \texttt{E-Q5-Macro-F1} compared to the sequence architecture (excluding teams with unidentified methods). However, in terms of results on the CEE task, pipeline models are leading in the \texttt{E-CEE-Macro-F1} score. Although our model's \texttt{E-Q5-Macro-F1} score is more than $8\%$ lower than the winning team, when compared with the same backbone model, \texttt{VietAI/t5-base}, we are only $1.5\%$ lower than them.

\begin{table*}[!ht]
    \centering
    \caption{Comparison of F1 results between MTP-COQE, top-performing teams, and baseline results provided by the organizers. Model approaches are categorized as \textit{E} for end-to-end, \textit{P} for pipeline, \textit{H} for hybrid methods, and $\_$ for unidentified approaches. Blue rows highlight our results, with the highest score in each column bolded.}
    \begin{adjustbox}{width=0.75\linewidth}
    \begin{tabular}{c l|r r r r r }
    \hline
        \textbf{Approach} & \textbf{Method} & \textbf{E-Q5-Macro} & \textbf{E-Q5-Micro} & \textbf{E-Q4} & \textbf{E-CEE-Macro} & \textbf{E-CEE-Micro}  \\ \hline
        E & ABCD team\textsubscript{viT5-large} & \textbf{23.73} & \textbf{29.52} & 30.29 & 64.11 & 63.31 \\ \hline
        \_ & NOWJ 3 & 23.00 & 26.84 & 29.07 & 60.79 & 59.89  \\ \hline
        \_ & VBD\_NLP & 21.31 & 29.41 & \textbf{31.72} & 63.37 & 62.84  \\ \hline
        E & ABCD team\textsubscript{viT5-base} & {16.82} & \_ & \_ & \_& \_ \\ \hline
        \rowcolor{blue!10}
        E & MTP-COQE & 15.30 & 24.10 & 25.90 & 58.66 & 57.93  \\ \hline
        P & The challengers & 11.19 & 20.92 & 23.23 & \textbf{66.17} & \textbf{65.45}  \\ \hline
        \rowcolor{blue!10}
        E & T5 with CD & 11.15 & 19.00 & 20.42 & 55.57 & 54.62  \\ \hline
        \rowcolor{blue!10}
        E & T5-base & 10.44 & 18.84 & 19.94 & 54.50 & 53.67  \\ \hline
        E & \textit{VCOM\_generative} & \textit{9.23} & \textit{18.40} & \textit{21.53} & \textit{56.17} & \textit{55.51}  \\ \hline
        \rowcolor{blue!10}
        P & MS\textsubscript{BERT} & 8.39 & 15.38 & 20.19 & 59.43 & 58.92  \\ \hline
        P & \textit{VCOM\_pipeline } & \textit{6.68} & \textit{13.37} & \textit{16.96} & \textit{60.83} & \textit{60.11}  \\ \hline
        \rowcolor{blue!10}
         H & {gpt3.5-turbo} & {3.40} & {6.51} & {9.38} & {39.35} & {38.14}  \\ \hline
    \end{tabular}
    \end{adjustbox}
    
    \label{tab:vcom-result}
\end{table*}

Additionally, it is crucial to assess the performance of encoder-only models, which are a type of generative model like those from OpenAI and other large language models (LLMs). Although OpenAI models are trained on vast amounts of data, reaching up to hundreds of terabytes, their performance on COQE tasks is very low (as shown in Table \ref{tab:camera-coqe-final-result} and \ref{tab:vcom-result}). This highlights the importance of constructing a sequence-to-sequence (seq2seq) model and training it to learn the relationship between input and output before applying it to specific problems.

\subsection{Ablation Experiments}

In comparing the results of our ablation experimental findings on both datasets, \texttt{F1} score decreases gradually as components are removed. This demonstrates the effectiveness of the proposed architecture for the COQE task. The experimental results significantly decrease when removing the multi-perspective data augmentation component.

\subsection{Error Analysis}
The model performance on Vietnamese dataset is suboptimal. To gain deeper insights into the reasons behind these errors, we conduct a detailed analysis of the results for tasks CEE and CPC, shown in Table \ref{exact-CEE}, \ref{exact-CPC}, respectively. 

\begin{table}[!h]
    \centering
    \caption{Exact matching results for CEE task.}
    \begin{adjustbox}{width=\linewidth}
    \begin{tabular}{l|r r r r r r|r r }
    \hline
        ~ & \multicolumn{6}{|c|}{\textbf{CEE - Exact matching - F1}} & \multicolumn{1}{|c}{\textbf{Quadruple}} \\ \hline
        
        ~ & \textbf{Subject} & \textbf{Object} & \textbf{Aspect} & \textbf{Predicate} & \textbf{Macro} & \textbf{Micro} & \textbf{Q4}   \\ \hline
        MS\textsubscript{BERT} & \textbf{63.47} & \textbf{65.45} & \textbf{53.60} & 55.20 & \textbf{59.43} & \textbf{58.69} & 8.39  \\ \hline
        T5-base & 58.73 & 60.75 & 46.90 & 51.64 & 54.50 & 53.67 & 10.44  \\ \hline
        T5 with CD & 61.36 & 62.61 & 46.92 & 51.38 & 55.57 & 54.62 & 20.42  \\ \hline
        \textbf{MTP-COQE} & 62.75 & 63.62 & 53.02 & \textbf{55.24} & 58.66 & 57.93 & \textbf{25.90}  \\ \hline
    \end{tabular}
    \end{adjustbox}
    \label{exact-CEE}
\end{table}
For the CEE task, our reproduced \textbf{MS\textsubscript{BERT}} model achieved the highest \texttt{E-CEE-Macro-F1} score, outperforming our generative base model by $5\%$ but \textbf{MTP-COQE} by only $0.76\%$. However, when evaluating quadruple extraction result \texttt{E-Q4-F1}, \textbf{MS\textsubscript{BERT}} performed significantly worse than the generative model, achieving only one-third of the performance of the our \textbf{MTP-COQE} model. 

For CPC task, the generative model variants also produced relatively higher results. Our proposed model performs much better on the CEE task compared to the CPC task, indicating that the generation model is not yet fully optimized for concatenating the components of a quintuple.
\begin{table}[!h]
    \centering
    \caption{Exact matching results for CPC task.}
    \begin{adjustbox}{width=\linewidth}
    \begin{tabular}{l|| r r r r r r r r r }
    \hline
        ~ & \multicolumn{9}{|c}{\textbf{CPC - Exact matching - F1}}  \\ \hline
        ~ & \textbf{COM} & \textbf{COM-} & \textbf{COM+} & \textbf{SUP} & \textbf{SUP-} & \textbf{SUP+} & \textbf{EQL} & \textbf{DIF} & \textbf{Macro} \\ \hline
        MS\textsubscript{BERT} & 0 & 13.72 & 19.44 & 0 & 0 & 13.24 & 13.68 & 7.02 & 8.01 \\ \hline
        T5-base & 0 & {21.93} & \textbf{22.91} & 0 & 0 & 10.53 & 16.48 & 11.67 & 10.44  \\ \hline
        T5 with CD & 0 & 16.67 & 21.90 & 0 & 0 & \textbf{14.29} & 20.28 & 8.93  & 10.26\\ \hline
       \textbf{MTP-COQE} & 0 & \textbf{31.66} & 20.18 & 0 & 0 & 11.85 & \textbf{21.74} & \textbf{12.00} & \textbf{12.18} \\ \hline
    \end{tabular}
    \end{adjustbox}
    \label{exact-CPC}
\end{table}

\begin{figure}
    \centering
    \includegraphics[width=0.85\linewidth]{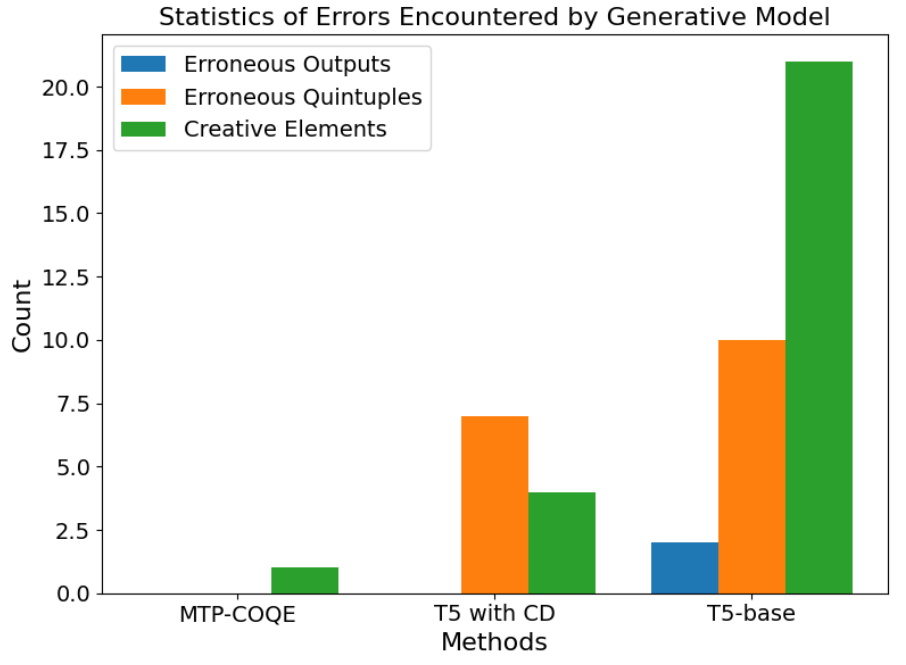}
    \caption{Statistics of errors encountered by the generative model.}
    \label{fig:error-table}
\end{figure}

As mentioned, a drawback of generative models is their dependence on learned knowledge, making it unpredictable what output the model will generate. Figure \ref{fig:error-table} summarizes errors observed on the Camera-COQE dataset, including nonsensical output sequences, deviations from the quintuple format (incorrect order or missing elements), and spontaneously generated creative elements not present in the original sentence.

\section{Conclusion}
 This paper introduces MTP-COQE, a novel method that combines generative models with multi-view prompt-based learning to achieve superior performance in extracting quintuples from product reviews across English and Vietnamese datasets. Our approach demonstrates significant advancements in comparative opinion mining tasks. However, challenges such as the inherent opacity of generative models, limited context awareness, and high computational costs for model refinement need consideration. Future research can explore integrating external knowledge, enhancing contextual understanding mechanisms, and developing modular frameworks for improved model control. 



\bibliographystyle{IEEEtranN}
\bibliography{references}

\end{document}